\documentclass[9pt,twocolumn,twoside]{pnas-new}

\templatetype{pnasresearcharticle} 

\title{A Fast Algorithm for Clustering of High Dimensional Feature Vectors}

\author[a,1]{Shahina Rahman}
\author[a,1,2]{Valen E. Johnson}

\affil[a]{Texas A\&M University}

\leadauthor{Rahman} 

\significancestatement{Clustering algorithms are used to separate objects into groups based on the similarity of features measured on each object. Clustering algorithms are used in many fields, including genomics, computer vision, machine learning, and statistical analyses. Nonetheless, clustering items based on high dimensional feature vectors remains a challenging unsupervised learning task, and existing algorithms are often unable to reliably identify meaningful clusters. We propose a model-based clustering algorithm that is insensitive to both the number and distributions of features that are measured on each object.  The algorithm does not require tuning parameters, is computationally faster than competing algorithms, and provides accurate estimates of cluster configurations in a benchmark group of gene expression datasets.}

\authorcontributions{Both authors contributed to the development of methodology, manuscript preparation and algorithm development. Rahman developed R code to implement the clustering algorithm.}
\authordeclaration{The authors declare no conflict of interest.}
\equalauthors{\textsuperscript{1}Shahina Rahman and Valen E. Johnson contributed equally to this work }
\correspondingauthor{\textsuperscript{2}To whom correspondence should be addressed. E-mail: vjohnson@stat.tamu.edu}


\begin{abstract}
We propose an algorithm for clustering high dimensional data. 
If $P$ features for $N$ objects are represented in an $N\times P$ matrix ${\bf X}$, where $N\ll P$, the method is based on exploiting the cluster-dependent structure of the $N\times N$ matrix ${\bf  XX}^T$.  Computational burden thus depends primarily on $N$, the number of objects to be clustered, rather than $P$, the number of features that are measured. This makes the method particularly useful in high dimensional settings, where it is substantially faster than a number of other popular clustering algorithms. Aside from an upper bound on the number of potential clusters, the method is independent of tuning parameters. When compared to $16$ other clustering algorithms on $32$ genomic datasets with gold standards, we show that it provides the most accurate cluster configuration more than twice as often than its closest competitors. We illustrate the method on data taken from highly cited genomic studies.
\end{abstract}

\dates{This manuscript was compiled on \today}

\usepackage{geometry}                		
\usepackage{graphicx}
\usepackage{subfigure}						
\usepackage{amsmath, bm}
\usepackage{graphicx}
\usepackage{amssymb}
\usepackage{epstopdf}
\usepackage{lscape}

\newcommand{\xkm}{x_{km}}

\newcommand{\bfX}{{\bf X}}
\newcommand{\bX}{{\bf X}}

\newcommand{\EJkm}{\mbox{E}(J_{km})}
\newcommand{\Ekm}{\mbox{E}(X_{km})}
\newcommand{\Elm}{\mbox{E}(X_{lm})}
\newcommand{\EJlm}{\mbox{E}(J_{lm})}
\newcommand{\Var}{\mbox{Var}}
\newcommand{\Cov}{\mbox{Cov}}
\newcommand{\E}{\mbox{E}}

\newcommand{\bJ}{{\bf J}}

\newcommand{\bG}{{\bf R}}

\newcommand{\bfmu}{{\boldsymbol \mu}}
\newcommand{\bfS}{{\boldsymbol \Sigma}}

\usepackage{color}
\definecolor{maroon}{cmyk}{0,0.87,0.68,0.32}

\begin{document}

\maketitle
\thispagestyle{firststyle}
\ifthenelse{\boolean{shortarticle}}{\ifthenelse{\boolean{singlecolumn}}{\abscontentformatted}{\abscontent}}{}

\bigskip

\dropcap{C}lustering algorithms are used to separate objects into groups, called clusters, based on the similarity of their features. Clustering algorithms have broad application throughout science, including most disciplines in the life and physical sciences. Despite their ubiquity, clustering of objects based on high dimensional feature vectors remains a challenging unsupervised learning task, which is further complicated by the fact that the number of clusters is seldom known a priori. Indeed, this task has become even more challenging as the number of features that can be measured on each object has grown rapidly with advances in technology. 

To address this challenge, we propose an algorithm for clustering high dimensional data in which computational burden is largely independent of the number and distribution of features measured on each item. Instead, computational burden depends primarily on the number of objects to be clustered. We compare the performance of this method with $16$ other clustering algorithms using $32$ genomic datasets for which a gold standard is available, and find that the proposed method provides the most accurate cluster configuration more than twice as often as its closest competitors.

\section*{Method}
The goal of our clustering algorithm is to partition $N$ objects into $C$ clusters based on the measurement of $P$ features on each object. The number of clusters $C$ is not assumed to be known in advance. We assume that the data are complete, meaning that all features are measured on each object. We organize the data into a $N\times P$ matrix ${\bX}$ in which the measurement in row $k$, column $m$ of ${\bfX}$, denoted by $\xkm$, represents the measurement of feature $m$ on object $k$. Letting $\delta_k$ denote the cluster membership of item $k$, $\delta_k \in \{ 1,\dots,C\}$, we assume a finite mixture model of the form  \[ x_{np} \sim F_{\delta_n,p},\quad n=1,\dots,N, \quad p=1,\dots,P, \]
where $F_{c,p}$ are arbitrary distributions assumed to satisfy $E(x_{np}^4)  < M$ for some fixed $M$, and $x_{np}$ is independent of $x_{m,q}$ if $n\neq m$ or $p\neq q$. If $\delta_k=\delta_l$, we write $k \simeq l$; otherwise $k\not\simeq l$.

With this notation, the clustering algorithm can be described by the following three steps:
\begin{enumerate}
\item Construct the $N\times N$ matrix $\bG=\bX\bX^T/P$. 
\item Juxtapose the elements of $\bG$ to construct the $N\times (N+1)$ matrix $\bJ$ in which $\EJkm=\EJlm$ for $m=1,\dots,N+1$ when $k \simeq l$ and $k,l \neq m$.
\item Apply standard clustering techniques to the $(N+1)\times 1$ vectors defined as the rows of $\bJ$.  
\end{enumerate}

In Step 1, we assume that the expected value of $\xkm$, $\Ekm$, differs from $\Elm$ for at least some features whenever $k \not\simeq l$.   In contrast, if $k\simeq l$, we assume that $E(X_{km})=E(X_{lm})$ for every feature $m$.  Based on these assumptions, the structure of $\bfX$ is described by the following lemmas.

{\bf Lemma 1.}  Suppose $k$, $l$, $m$ and $n$ represent indices of distinct items. Then the following relations hold:
\begin{enumerate}
\item If $k\simeq l$, then $\E(R_{kk}) = \E(R_{ll})$,  
\item If $k\simeq l \simeq m \simeq n$, then $\E(R_{km})=\E(R_{kl} )=\E(R_{ml}) = \E(R_{ln} )$. 
\end{enumerate}

{\bf Lemma 2.} 
Define $\theta_{\delta_n p} = E(X_{np})$.  If $\theta_{cp}$, $c=1,\dots,C$ and $p=1,\dots,P$, are drawn independently from a distribution with a continuous density with respect to Lebesque measure, then with probability equal to 1: 
\begin{enumerate} 
\item If $k \not\simeq m$, then $\E(R_{kk} )\neq \E(R_{mm})$,
\item If $k \not\simeq r$ or $l \not\simeq s$, and if $l\not\simeq r$ or $k\not\simeq s$, then $\E(R_{kl}) \neq \E(R_{rs})$ .
\end{enumerate}

Except in degenerate situations, statements 1 and 2 of Lemma 2 will generally hold, although the difference between expectations may become small if the proportion of non-informative features approaches 1.

A visual representation of these relations can be obtained if the rows of $\bX$ (and thus $\bG$) are arranged according to cluster. Assuming this to be the case, Figure 1 depicts the structure of $\bG$ when there are three clusters.  In this figure, $\mu_c$ denotes the expected value of the diagonal elements of $\bG$ corresponding to cluster $c$ and $\mu_{cd} \, (=\mu_{dc})$ the expected value of off-diagonal elements in which the corresponding objects belong to clusters $c$ and $d$.

Having formed the matrix $\bG$, in Step 2 we juxtapose its entries to construct the $N\times (N+1)$ matrix $\bJ = \{ J_{kl}\}$.  For $k,l\leq N$ and $k\neq l$, the elements of $\bJ$ are defined as follows: 
\[ J_{kl} = R_{kl}, \quad J_{k,N+1}=R_{k,k}, \quad J_{kk} = \sum_{l\neq k} \frac{R_{kl}}{N-1} .\] 

For example with $N=3$, this construction produces a juxtaposed matrix $\bJ$ given by 
\[ \bJ = 
\begin{bmatrix}
\frac{R_{21}+R_{31}}{2} & R_{12} & R_{13} & R_{11} \\
R_{21} & \frac{R_{12}+R_{32}}{2} & R_{23} & R_{22} \\
R_{31} & R_{32} & \frac{R_{13}+R_{23}}{2} & R_{33} 
\end{bmatrix} .
\]



\medskip

The structure of $\bf J$ has several properties that facilitate model-based clustering.  With the exception of the diagonal entries, $E(J_{km} )= E(J_{lm})$ whenever objects $k$ and $l$ are in the same cluster. The violation of this equality for the diagonal elements only affects one entry in each row of $\bJ$, and does not significantly affect the outcome of standard clustering algorithms applied to the rows of $\bJ$ when $N$ is moderately large.

The construction of the $\bJ$ matrix also results in a simple structure on the covariance between elements within any row. Namely, the covariance between any pair of elements $l$ and $m$ in row $k$ depends only on the cluster membership of $k$, $l$, and $m$.  Hence the number of parameters that must be estimated to define the covariance matrices for the rows of $\bJ$ is a function $C$ (not $N$), and equals 
$4\binom{C}{1} + 3\binom{C}{2} + \binom{C}{3}$, which is ${ O}(C^3)$. The following lemma describes the mean and covariance structure of $\bJ$.  

{\bf Lemma 3.} Define $\theta_{\delta_n p}$ as above, $\varsigma_{\delta_n p} = E(X_{np}^2)$, $\gamma_{\delta_n p} = E(X_{np}^3)$, and $\kappa_{\delta_n p} = E(X_{np}^4)$.  Then the following hold for all $k,m,n \leq N$ (all sums range from $p=1,\dots,P$ unless otherwise indicated):
\begin{enumerate}
\item $ \E(J_{k,N+1}) = \sum \varsigma_{\delta_k p}/P \equiv \mu_{\delta_k}$
\item For $k\neq m$, $\E(J_{km}) = \sum \theta_{\delta_k p} \theta_{\delta_m p}/P \equiv \mu_{\delta_k\delta_m}$
\item \[ \E(J_{kk}) = \frac{1}{(N-1)P} \sum_{\substack{m=1 \\ m\neq k}}^N \E(J_{km}) = \frac{1}{(N-1)P} \sum_{\substack{m=1 \\ m\neq k}}^N \mu_{\delta_k\delta_m}\] 
\item $\Var(J_{k,N+1}) = \sum (\kappa_{\delta_k p} - \varsigma_{\delta_k p}^2)/P^2 \equiv \sigma^2_{\delta_k} $
\item For $k\neq m$, \newline
$\Var(J_{km}) = \sum (\varsigma_{\delta_k p} \varsigma_{\delta_m p}- \theta_{\delta_k p} \theta_{\delta_m p})/P^2 \equiv \sigma^2_{\delta_k\delta_m}$
\item For $m,n\neq k$, $m\neq n$, \newline $\Cov(J_{km},J_{kn}) = \sum (\varsigma_{\delta_k p}-\theta_{\delta_k p}^2) \theta_{\delta_m p}\theta_{\delta_n p}/P^2 \equiv \sigma^2_{\delta_k\delta_m\delta_n}$
\item 
\begin{eqnarray*} 
\Var(J_{kk})  
&=& \frac{1}{(N-1)^2}  \sum_{\substack{m=1 \\ m\neq k}}^N \sigma^2_{\delta_k\delta_m} \\ 
& +&  \frac{2}{(N-1)^2} \sum_{\substack{m=1 \\ m\neq k}}^{N-1} \sum_{\substack{l=m+1 \\ l \neq k }}^N \sigma^2_{\delta_k\delta_m\delta_l}
\end{eqnarray*}
\item For $k \neq m$, \newline
\[ \Cov(J_{kk},J_{km})  = \frac{1}{N-1}\sum_{\substack{n=1 \\n\neq k}}\sigma^2_{\delta_k\delta_m\delta_n}\]
\end{enumerate}
Figure 2 depicts the structure of the covariance matrix of the rows of $\bJ$ when the number of underlying clusters equals 2.  In this figure, items corresponding to rows 1-3 of $\bJ$ are assumed to fall into cluster 1 and items corresponding to rows 4-6 into cluster 2.  The covariance matrices corresponding to rows 1 and 5 are displayed.  The values indicated by a ``$\cdot$'' are determined by statements 7 and 8 of Lemma 3. 
We now describe the asymptotic distribution of the rows of $\bJ$.  


{\bf Lemma 4.} Under certain regularity condition, ${\bG}_k$ denote row $k$ of $\bG$, and define $\bfmu^*_{\delta_k} = \{ E(R_{km}) \}_{m=1}^{N}$ and $\bfS^*_{\delta_k} = \{ \Cov(R_{km},R_{kn})  \}_{m,n=1}^{N}$. If $N$ is fixed and $P\rightarrow \infty$, then
\[ \bG_k \stackrel{d}{\longrightarrow} {\bf MVN}_{(N+1)}(\bfmu^*_{\delta_k},\bfS^*_{\delta_k}) .\]
That is, $\bG_k$ converges in distribution to a degenerated $(N)$ dimensional multivariate normal distribution with mean $\bfmu_{\delta_k}$ and covariance matrix $\bfS_{\delta_k}$. 

Following Lemma 4, it applies to the $k^{th}$ row of \bJ, noting the diagonal elements are a linear combination of the remaining elements of the row. In Lemma 3, we had defined $\bfmu_{\delta_k} = \{ E(J_{km}) \}_{m=1}^{N+1}$ and $\bfS_{\delta_k} = \{ \Cov(J_{km},J_{kn})  \}_{m,n=1}^{N+1}$. Given $C$ (but not $\delta_k$), it follows from Lemma 4 that the marginal density of $\bJ_k$ can be approximated by a C-component degenerated Gaussian mixture 
\[
f(\bm{J_k})  = \sum_{a =1}^Cw_a\phi(\bJ_k;\bm{\mu_a},\bm{\Sigma_a}) ,
\]
where the mixing weights $w_a$ satisfy $w_a \geq 0$ for all $1 \leq a \leq C$ and $\sum_{a = 1}^Cw_a = 1$, and $\phi(J_k;\bm{\mu_a},\bm{\Sigma_a}) $ denotes the multivariate Gaussian density function with mean vector $\bm{\mu_a}$  and covariance matrix ${\bm\Sigma_c}$, 
\[ 
\frac{1}{(2\pi)^{p/2}\hbox{det}(\bm{\Sigma_a})^{1/2}}\times \hbox{exp} \big\{ - \frac{1}{2}(\bJ_k - \bm{\mu_a}){\bm{\Sigma_a}}^{-1}(\bJ_k - \bm{\mu_a})^{T}\big\}. 
\]

In Step 3 of our algorithm, we treat the $N$ rows of $\bJ$ as if they were independent normal random variables and apply standard model-based clustering algorithms to them. We refer to the resulting algorithm as R-J clustering. We note that the rows of $\bJ$ are not actually independent because of the symmetry of ${\bf R}$. This assumption is examined further in the Discussion.

Because of the pervasive scientific need to cluster objects, there are a broad range of clustering algorithms that might be chosen to cluster the rows of {\bf J}.  Excellent reviews of such algorithms are provided in \cite{aggarwal2013data,daxin2004cluster}; readers interested in statistical model-based clustering algorithms might also consult, for example, \cite{booth2008clustering,hoff2006model,mclachlan1988mixture,quintana2003bayesian,rousseau2011asymptotic}. For our purposes, the {\it mclust} algorithm developed in \cite{banfield1993model,fraley1998many,fraley1999mclust,fraley2002model} is most relevant because it is a model-based clustering algorithm based on an underlying assumption that the feature vectors in each cluster follow a common multivariate normal distribution. 



Unfortunately, the {\em mclust} algorithm cannot be used to exactly optimize the R-J clustering algorithm because the {\em mclust} algorithm constrains the underlying covariance matrices $\{ \Sigma_c\}$ to be diagonal matrices in $N\leq P$ settings.
Nonetheless, for each value of $C$ between $1$ and a user-specified maximum ${C}_{max}$, we use the {\em mclust} algorithm to obtain an initial estimate of the cluster configuration, and then apply an estimation-maximization (EM) algorithm \citep{dempster1977maximum} to maximize the mixture likelihood function associated with the $\bJ$ matrix for that value of $C$. The BIC criterion is then used to choose the best model. The specific steps in the model may be summarized as follows:

\medskip

\noindent{\bf Computational Strategy:}\newline

\noindent For $C =1,\dots,C_{max}$
\begin{enumerate}
\item Apply {\em mclust algorithm}:
\begin{enumerate}
\item Obtain the initial cluster configuration with $C$ clusters using agglomerative clustering.
\item Perform the EM algorithm with diagonal covariance matrices to obtain a local maximum (i.e., ``VVI'' covariance specification in {\em mclust}).
\item If any cluster shrinks to one item, set $C_{max} = C-1$ and exit loop. 
\end{enumerate}
\item Using the configuration obtained in 1b, perform the EM algorithm based on the exact mean and covariance structure.
\begin{enumerate}
\item E-step: Compute posterior probability item $k$ belongs to each cluster $a$, $1\leq a \leq C$: 
\begin{eqnarray*}
Q^k_a &=& P(\delta_k = a \vert \bJ_{k};\bm{\widehat{w}_a}, \bm{\widehat{\mu}_a}, \bm{\widehat{\Sigma}_a}) \\
&=& \dfrac{\bm{\widehat{w}_a}\phi(\bJ_k;\bm{\widehat{\mu}_a},\bm{\widehat{\Sigma}_a})}{\sum_{a = 1}^c\bm{\widehat{w}_a}\phi(\bJ_k;\bm{\widehat{\mu}_a},\bm{\widehat{\Sigma}_a})}.
\end{eqnarray*}
\item M-step: Re-estimate mixture parameters using current values of $Q^k_a$: \newline
For $1\leq a,b,d \leq C$
\begin{enumerate}
\item $\widehat{w}_a  = \sum_{k = 1}^N Q^k_a/N$,
\item $\widehat{\mu}_{a} = \sum_{k=1}^N Q^k_a J_{kk}/\sum_{k=1}^n Q^k_a$,
\item $\widehat{\mu}_{ab} = \sum_{\substack{k,m \\ k\neq m}}^N Q^k_aQ^m_b J_{km}/\sum_{\substack{k,m \\ k\neq m}}^N Q^k_aQ^m_b$,
\item $\widehat{\sigma}^2_{a} =  \sum_{k=1}^N Q^k_a (J_{kk} - \widehat{\mu}_{a0})^2/\sum_{k=1}^N Q^k_a$, 
\item \[\widehat{\sigma}^2_{ab} =  \sum_{\substack{k,m \\ k\neq m }}^N Q^k_aQ^m_b (J_{km}- \widehat{\mu}_{ab})^2/\sum_{\substack{k,m \\ k\neq m}}^N Q^k_aQ^m_b,\]
\item \[\widehat{\sigma}^2_{abd} =  \frac{\sum_{\substack{k,m,l \\ k\neq m \neq l }}^N Q^k_aQ^m_bQ^l_d (J_{km}- \widehat{\mu}_{ab})(J_{kl}- \widehat{\mu}_{ad})}{\sum_{\substack{k,m,l \\ k\neq m \neq l}}^N Q^k_aQ^m_bQ^l_d},\]
\end{enumerate}
\end{enumerate}
\item Record maximized log-likelihood, $L_C(\widehat{\bm{w}}, \widehat{\bm{\mu}}, \widehat{\bm{\Sigma}})$.
\end{enumerate}
Select the optimal cluster configuration based on the BIC criterion, defined as the configuration that maximizes $2L_C(\widehat{\bm{w}}, \widehat{\bm{\mu}}, \widehat{\bm{\Sigma}}) - M log(N)$, where $M$ is the number of mixing parameters in the model that produced $L_C$.

\medskip

Further details and code to implement this strategy are provided in the Supplemental Materials.

\section*{Results}

Shah and Koltun \cite{shah2017robust} provide a recent comparison of 14 clustering methods for 32 gene expression datasets based on adjusted mutual information (AMI).  The AMI between two cluster configurations, say $A$ and $B$, is defined as 
\[ \hbox{AMI}(A,B) = \dfrac{MI(A,B) - E[MI(A,B)]}{\sqrt{H(A)H(B)}- E[MI(A,B)]}, \] 
where $H(\cdot)$ denotes entropy and MI($\cdot, \cdot$) denotes mutual information. An AMI value of 1 occurs when the two partitions are equal, and 0 represents the AMI value expected by chance under a hypergeometric sampling model for the partitions.

In the Shah and Koltun comparisons, validated cluster configurations were available for all $32$ gene expression datasets. The 14 methods included in their comparison were k-means++ \cite{arthur2007k}, Gaussian mixture models (GMM), fuzzy clustering, mean-shift clustering (MS) \cite{comaniciu2002mean}, two variants of agglomerative hierarchical clustering (HC-Complete and HC-Ward), normalized cuts (N-Cuts) \cite{shi2000normalized}, affinity propagation (AP) \cite{frey2007clustering}, Zeta l -links (Zell) \cite{zhao2009cyclizing}, spectral embedded clustering (SEC) \cite{nie2009spectral}, clustering using local discriminant models and global integration (LDMGI) \cite{yang2010image}, graph degree linkage (GDL) \cite{zhang2012graph}, path integral clustering (PIC) \cite{zhang2013agglomerative}, robust continuous clustering (RCC) and robust clustering with dimension reduction (RCC-DR) \cite{shah2017robust}. The parameter settings for these methods are summarized in the supplementary material of \cite{shah2017robust}. In addition, we also included unsupervised clustering methods that are widely applied to genomic data sets, the method of GAP statistics \cite{tibshirani2001estimating} implemented with partitioning around mediods, and consensus clustering \cite{monti2003consensus} with k-means and hierarchical complete linkage baselines. The parameter settings used for these methods are provided in Table S5 in the supplementary materials. 

The datasets used for the comparisons are available at
\href{https://schlieplab.org/Static/Supplements/CompCancer/datasets.htm}{https://schlieplab.org/Static/Supplements/CompCancer\\/datasets.htm}. The gene filtering schemes used for these data are also available from this site and are discussed in \cite{de2008clustering}. Before applying R-J clustering to these datasets, we applied a standard transformation to the feature space (i.e., gene expression values): We took the logarithm of the gene expression values, subtracted the median of each gene's logged expression value across subjects, and divided by the standard deviation of the log-transformed values. For data sets that were already preprocessed and contained negative transformed values, we analyzed the pre-processed data without further transformations.  (A large-scale study on the impact of these transformation procedures for cancer gene expression data has been recently presented in \cite{de2008comparative} and suggests that appropriate transformations on the feature space are required to produce meaningful analyses.)  Further details on which datasets were transformed appear in the Supplementary Material. 

Table \ref{AMI} provides the AMI values reported in \cite{shah2017robust}, supplemented with the AMI values obtained for R-J clustering. For each dataset, the method that achieved the highest AMI is highlighted in blue. In $12$ of the $32$ datasets, R-J clustering produced the highest AMI value. SEC and RCC-DR were ranked second in this comparison, with each achieving the highest AMI value in $5$ datasets. Fuzzy clustering and mean shift clustering results are not displayed due to formatting constraints and the fact that these algorithms did not provide optimal AMI values for any dataset. Note that the kmeans++, GMM, AC-W, N-Cuts and SEC algroithms require pre-specification of the number of clusters.  

Because the R-J clustering algorithm is applied to a $N\times N$ matrix rather than a $N\times P$ matrix, the first step in its estimation algorithm (i.e., application of the {\em mclust} algorithm) tends to be faster than other model-based clustering algorithms.  To illustrate this fact, in Table \ref{Comp_times} we display execution times of the {\em mclust} algorithm applied to the {\bf J} matrix, along with the execution times of affinity propagation (AP), mean shift clustering (MS), GAP statistics, consensus clustering with k-means and hierarchical clustering baselines (CC-km and CC-hc), and robust continuous clustering (RCC and RCC-DR) for each of the $32$ datasets from Shah and Koltun.  The other methods in Table~1 were not included in this comparison because those methods required the prior specification of the number of clusters $C$. From Table~2 we see that the {\em mclust} algorithm applied to the {\bf J} matrix and AP clustering methods required median execution times of $0.09$ and $0.06$ seconds. The median time required by R-J clustering implemented with the full covariance matrix was $5.86$ seconds.  This execution time was obtained using a non-optimized R program \cite{R} (see Supplemental Materials).  Unfortunately, the mean execution time for the R-J clustering algorithm increases rapidly with $N$, as witnessed by the relatively long execution times observed for the Gordon, Ramaswamy, Su, and Yeoh \cite{gordon2002translation, ramaswamy2001multiclass, su2001molecular, yeoh2002classification} datasets for which $N=181, 190, 174$ and $248$, respectively. We anticipate that this deficiency can be overcome by consensus-clustering type applications of the R-J algorithm in large $N$ settings.  In comparison, RCC and RCC-DR required median execution times of $14.57$ and $1.29$ seconds, and GAP and consensus clustering required median execution times of $8.57$ and $8.23$ seconds, respectively. 
All code was run on a workstation with an Intel(R) Core(TM) i7-3770 CPU clocked at 3.40GHz with 8.00 GB RAM.  

Output from the R-J clustering algorithm includes the number of clusters, the cluster membership of each item, and the estimated mean vector and covariance matrix for each observation vector. These values can be conveniently visualized using heatmaps. More detailed application of the algorithm to the genomic datasets reported in \cite{dyrskjot2003identifying,chowdary2006prognostic} are illustrated in the Supplemental Material.

\section*{Discussion}
The R-J clustering algorithm offers a fast and comparatively accurate method for clustering objects based on high-dimensional feature vectors.  The algorithm assumes a finite mixture formulation on feature vectors in which mixing densities are assumed to have a bounded fourth moment. Measurements made on distinct features are assumed to be statistically independent. 

The objective function for the R-J clustering algorithm is motivated by the assumption that rows of the {\bf J} matrix are independent and have multivariate normal distributions.  Of course, the symmetry of the $\bJ$ matrix implies that this assumption is not satisfied. A more accurate statistical model could be obtained by modeling the upper triangle portion of the $\bG$ matrix directly, but doing so greatly enhances the computational complexity of resulting algorithms: the upper triangle of $\bG$ contains $N(N+1)/2$ elements, implying a precision matrix containing approximately $N^4/8$ elements.  Calculating likelihood functions based on this matrix would significantly increase the computational burden required to evaluate cluster configurations.

Application of the R-J clustering algorithm to a number of genomic datasets suggests that the algorithm provides useful clustering of samples into clusters.  The performance of the algorithm on simulated data generated from finite mixture models (e.g., \cite{guo2010pairwise}) often produces nearly optimal cluster configurations, although the connection between simulated and real data is often tenuous.  We expect that future developments of the R-J algorithm will facilitate its use for large $N$, large $P$ settings, and that it will be possible to develop useful diagnostics to identify those settings in which the R-J algorithm provides an adequate clustering of sampled items. 

\section*{Acknowledgments}
We thank Anirban Bhattacharya and Irina Gaynavova for helpful comments and discussions, and Marina Romanyuk for assistance in calculating computation times of various algorithms.  Both authors acknowledge support from NIH grant CA R01 158113.

\begin{figure}
\vspace{.05in}
\begin{center}
\includegraphics[scale = 0.4]{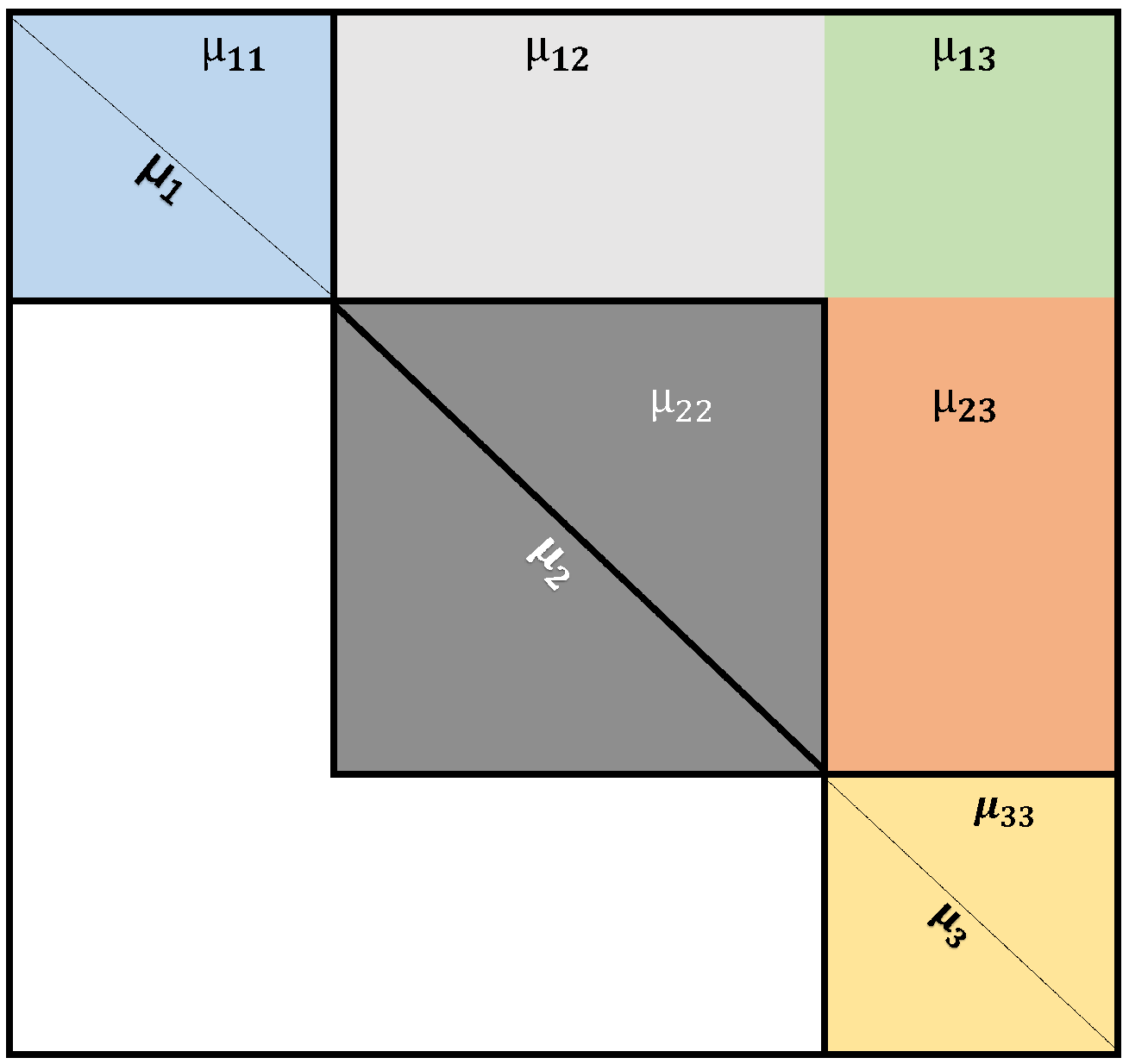} 
\end{center}
\caption{Mean structure of $\bG$ matrix based on 3 clusters} \label{GMat}
\end{figure}

\begin{figure*}
\begin{center}
\begin{tabular}{c c c}
(a) Cov($\bJ_1$) belonging to cluster 1, $\bm{\Sigma_1}$  &   & (b)Cov($\bJ_4$) belonging to cluster 2, $\bm{\Sigma_2}$ \\
\includegraphics[scale = 0.6]{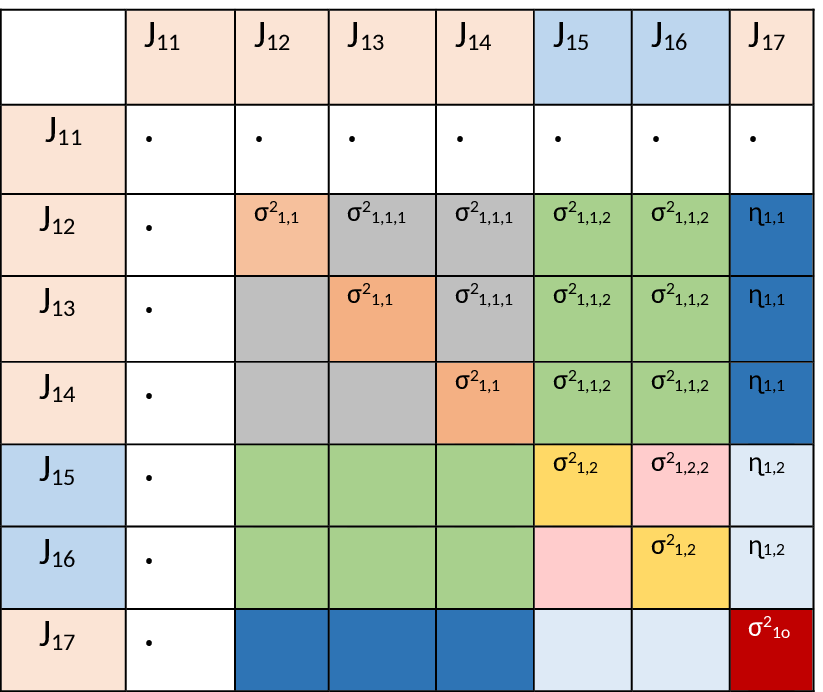} &  &\includegraphics[scale = 0.32]{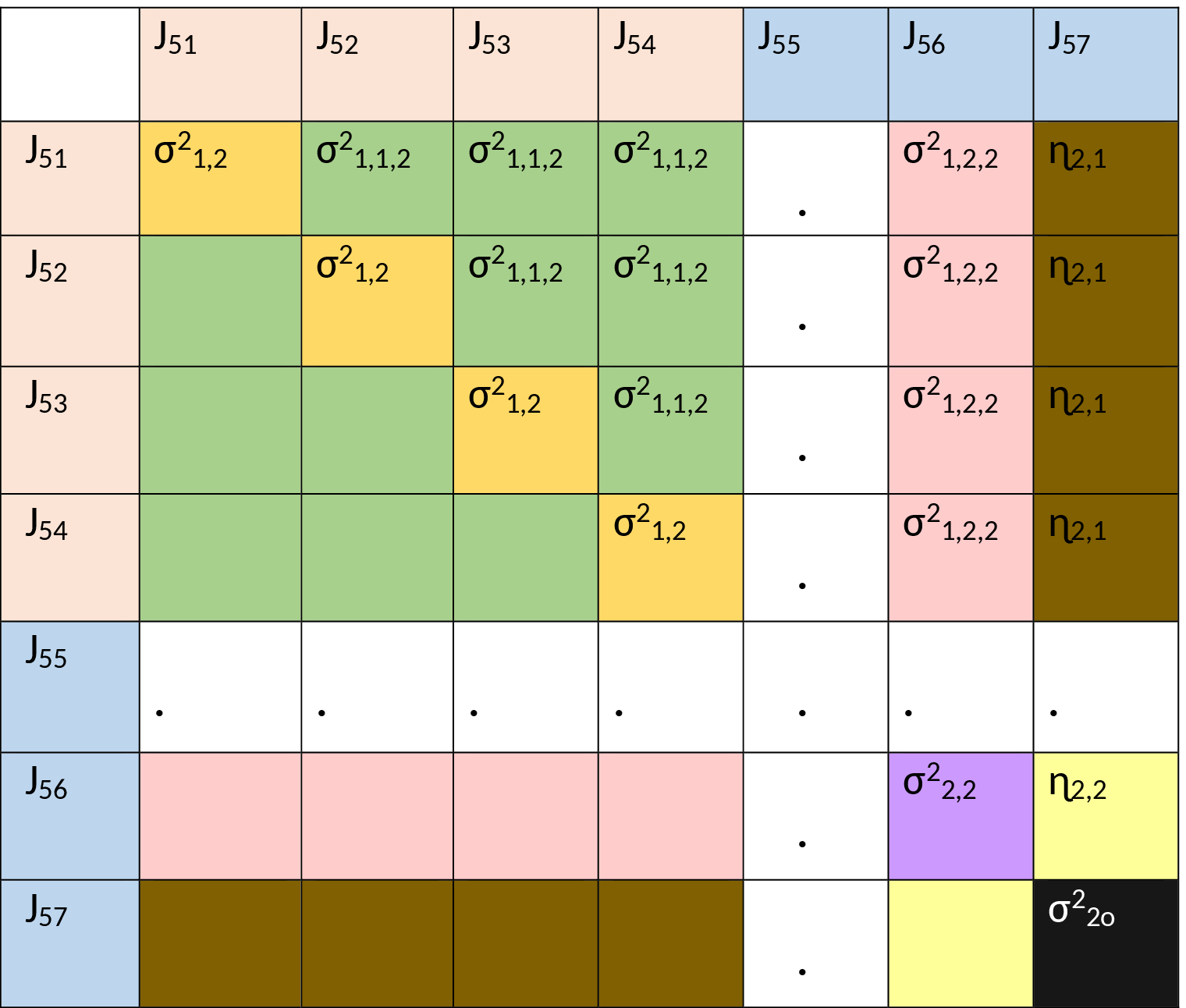} \\
\end{tabular}
\end{center}
\caption{The $(N+1)\times(N+1)$ covariance structure of $\bJ$ matrix based on $2$ clusters. The number of distinct covariance parameters that must be estimated in this example is $12$, and each value has been assigned a distinct color. The cells represented by dots correspond to covariances of the substituted diagonal elements; these values can be expressed as functions of the other parameters.}\label{JCov}
\end{figure*}

\begin{table*}
\caption{Adjusted Mutual Information for 15 clustering algorithms. For each dataset, the maximum achieved AMI is highlighted in blue.  R-J clustering achieves the highest AMI in 12 of the 32 datasets. Based on AMI values reported in \cite{shah2017robust}. The algorithms defined by kmeans++, GMM, AC-W, N-Cuts and SEC required pre-specification of the number of clusters.  Mean shift and fuzzy clustering results are not displayed due to formatting constraints; neither of these algorithms achieved the highest AMI for any dataset.} \label{AMI}
\resizebox{\textwidth}{!}{%
\begin{tabular}{ |>{\columncolor{maroon!15}}c|c|c|c|c|c|c|c|c|c|c|c|c|c|c|c| } 
\hline
\hline
\rowcolor{lightgray}
{\bf Datasets}  
&{\bf kmeans++}	&{\bf GMM}	&{\bf CC-km} &{\bf CC-hc} &{\bf GAP}	&{\bf AC-W}	&{\bf N-Cuts}	&{\bf AP}	&{\bf Zell}	&{\bf SEC}	&{\bf LDMGI}	&{\bf PIC}	&{\bf RCC}	&{\bf RCC-DR}  &{\bf RJ}\\ \hline \hline
Alizadeh-2000-v1    
&0.340	&0.024	&0.037	&0.007	&0.000	&0.101	&0.096	&0.232	&0.250&0.238 &0.123  &0.033 &0.000  &0.426  &\cellcolor{blue!30}0.515 \\ \hline
Alizadeh-2000-v2 
&0.568	&0.922	&0.695	&0.628	&\cellcolor{blue!30}1.000&0.922	&0.922	&0.563	&0.922 &0.922  &0.738	&0.922	&\cellcolor{blue!30}1.000	&\cellcolor{blue!30}1.000   &\cellcolor{blue!30}1.000   \\ \hline
Alizadeh-2000-v3
&0.586	&0.604	&0.551	&0.497	&0.629	&0.616	&0.601	&0.540	&0.702	&0.574 &0.582 &0.625 &\cellcolor{blue!30}0.792 &\cellcolor{blue!30}0.792  &\cellcolor{blue!30}0.792 \\ \hline
Armstrong-2002-v1
&0.372	&0.372	&0.478	&0.463	&0.475	&0.308	&0.372 &0.381 &0.308 &0.323 &0.355 &0.308  &0.528 &0.546  &\cellcolor{blue!30}0.547  \\ \hline
Armstrong-2002-v2 
&\cellcolor{blue!30}0.891	&0.803	&0.541	&0.375	&0.525 &0.746	&0.83 &0.586 &0.802 &\cellcolor{blue!30}0.891 &0.509  &0.802  &0.642 &0.838  &0.539 \\ \hline
Bhattacharjee-2001    
&0.444	&0.406	&0.598	&0.521	&0.518	&\cellcolor{blue!30}0.601	&0.563	&0.377	&0.496	&0.570  &0.378	&0.378	&0.495	&0.600   &0.557\\  \hline 
Bittner-2000    
&-0.012	&-0.002	&0.021	&0.024	&0.000	&0.002	&0.042	&\cellcolor{blue!30}0.243	&0.115	&-0.002	&0.014	&0.115	&-0.016	&0.156   &0.138\\  \hline 
Bredel-2005 
&0.297	&0.208	&0.202	&0.211	&0.035	&0.384	&0.203	&0.139	&0.278	&0.259	&0.295	&0.278	&\cellcolor{blue!30}0.468	&0.466   &0.265  \\  \hline
Chowdary-2006   
&0.764	&0.808	&0.499	&0.298	&0.000	&\cellcolor{blue!30}0.859	&\cellcolor{blue!30}0.859	&0.443	&\cellcolor{blue!30}0.859	&\cellcolor{blue!30}0.859	&\cellcolor{blue!30}0.859	&\cellcolor{blue!30}0.859	&0.360	&0.393   &0.585 \\  \hline
Dyrskjot-2003    
&0.507	&0.532	&0.236	&0.241	&0.348	&0.474	&0.303	&0.558	&0.269	&0.389	&0.385 &0.177	&0.359	&0.383   &\cellcolor{blue!30}0.623 \\  \hline
Garber-2001
&0.242	&0.137	&0.026	&0.026	&0.096	&0.210	&0.204	&\cellcolor{blue!30}0.274	&0.246	&0.200	&0.191	&0.246	&0.240	&0.173    &0.130  \\  \hline
Golub-1999-v1
&0.688&0.583	&0.688	&0.418	&0.044&\cellcolor{blue!30}0.831	&0.650&0.430	&0.615	&0.615	&0.615&0.615	&0.527	&0.490    &0.420  \\  \hline
Golub-1999-v2
&0.680	&0.730	&0.439	&0.282	&0.000	&\cellcolor{blue!30}0.737	&0.693	&0.516	&0.689	&0.703	&0.600	&0.689	&0.656	&0.597    &0.538  \\  \hline
Gordon-2002    
&0.651	&0.669	&0.651	&0.432	&0.435	&0.483	&0.681	&0.304	&-0.005	&\cellcolor{blue!30}0.791	&0.669	&0.664	&0.349	&0.343    &0.429\\  \hline
Laiho-2002
&0.007	&\cellcolor{blue!30}0.207&0.116	&0.044	&0.000	&-0.007	&0.030	&0.061	&0.073	&-0.007	&0.093	&0.044	&0.000	&0.000   &0.144\\  \hline
Lapointe-2004-v1 
&0.088	&0.141	&0.116 &0.147 &0.034 &0.151	&0.179	&0.162	&0.151	&0.088	&0.149	&0.151	&0.171	&0.156 &\cellcolor{blue!30}0.181 \\  \hline
Lapointe-2004-v2 
&0.008	&0.013	&0.092	&0.082	&0.199	&0.033	&0.153	&0.210	&0.147	&0.028	&0.118	&0.171	&0.155	&\cellcolor{blue!30}0.239  &0.172 \\  \hline
Liang-2005 
&0.301	&0.301	&0.236	&0.261	&0.243	&0.301	&0.301	&\cellcolor{blue!30}0.481 &0.301 &0.301	&0.301	&0.301	&0.401	&0.419  &\cellcolor{blue!30}0.481\\  \hline
Nutt-2003-v1
&0.171	&0.137	&0.219	&0.311	&0.000	&0.159	&0.156	&0.116	&0.109	&0.086	&0.078 &0.113	&0.142	&0.129  &\cellcolor{blue!30}0.425 \\  \hline
Nutt-2003-v2
&-0.025	&-0.025	&0.138	&0.131	&0.035	&-0.024	&-0.025	&-0.027	&-0.031	&-0.025	&-0.027	&-0.030	&-0.030	&-0.029 &\cellcolor{blue!30}0.435 \\  \hline
Nutt-2003-v3 
&0.063	&0.259	&0.163	&0.169	&0.000	&0.004	&0.080  &-0.002 &0.059  &0.080  &0.174 &0.059  &0.000  &0.000  &\cellcolor{blue!30}0.642 \\  \hline
Pomeroy-2002-v1
&0.012	&-0.022	&0.014 &0.007	&-0.007	&-0.020	&-0.006	&0.061	&-0.020	&0.008	&-0.026	&-0.020	&0.111 &\cellcolor{blue!30}0.140  &0.067 \\  \hline
Pomeroy-2002-v2 
&0.502	&0.544	&0.443	&0.309	&0.376	&0.591	&\cellcolor{blue!30}0.617&0.586	&0.568	&0.577	&0.602	&0.568	&0.582	&0.582  &0.246 \\  \hline
Ramaswamy-2001 
&0.618	&0.650	&0.189	&0.258	&0.336 &0.623	&0.651	&0.592	&0.618	&0.620	&0.663	&0.639	&0.635	&\cellcolor{blue!30}0.676  &0.613 \\  \hline
Risinger-2003 
&0.210	&0.194	&0.174	&0.152	&0.000
&0.297	&0.223  &0.309  &0.201 &0.258  &0.153  &0.201 &0.227  &0.248  &\cellcolor{blue!30}0.311\\  \hline
Shipp-2002-v1    
&\cellcolor{blue!30}0.264	&0.149	&0.079	&0.087	&0.079
&0.208	&0.132  &0.113  &-0.002 &0.168 &0.203  &-0.002 &0.134 &0.124 &0.065 \\ \hline
Singh-2002  
&0.048	&0.029	&0.032	&0.037	&0.066
&0.019	&0.033  &0.079 &-0.003 &0.069  &-0.003 &0.066  &0.034 &0.034    &\cellcolor{blue!30}0.159\\  \hline
Su-2001   
&0.666	&0.720	&0.426	&0.496	&0.589 &0.662	&\cellcolor{blue!30}0.738  &0.657 &0.687  &0.650 &0.667 &0.660 &0.725  &0.702  &0.622 \\  \hline
Tomlins  
&0.396	&0.366	&0.184	&0.196	&0.423
&0.454	&0.409	&0.374	&\cellcolor{blue!30}0.647&0.469	&0.419	&0.590	&0.485	&0.513 &0.459\\  \hline
Tomlins-2006-v2 
&0.368	&0.333	&0.172	&0.094	&0.000 &0.215	&0.292  &0.340	&0.226	&\cellcolor{blue!30}0.383	&0.354	&0.311	&0.348	&0.373   &0.294\\  \hline
West-2001  
&0.489	&0.413	&0.358	&0.337	&0.00 &0.489	&0.442	&0.258	&\cellcolor{blue!30}0.515	&0.489	&0.442	&\cellcolor{blue!30}0.515	&0.391	&0.391  &0.308 \\  \hline
Yeoh-2002-v2  
&0.385	&0.343	&0.021	&0.018	&0.000	&0.383	&0.479	&0.405	&0.530	&\cellcolor{blue!30}0.550	&0.337	&0.442	&0.496	&0.465   &0.127 \\  \hline
\hline 
\hline
\end{tabular}%
}
\end{table*}

\begin{table*}
\caption{Computation times for 8 clustering algorithms where the number of clusters are not required to be known apriori. } \label{Comp_times}
\resizebox{\textwidth}{!}{%
\begin{tabular}{ |>{\columncolor{maroon!10}}c|c|c|c|c|c|c|c|c|c| } 
\hline
\hline
\rowcolor{lightgray}
{\bf Datasets}  
&{\bf AP}	&{\bf GAP}	&{\bf CC-km} &{\bf CC-hc}	&{\bf RCC}	&{\bf RCC-DR}   &{\bf RJ-full} &{\bf RJ-mclust}\\ \hline \hline
Alizadeh-2000-v1    
&0.031		&3.82	&4.31	&3.08	&6.51	&0.97 &4.08 &0.08 \\ \hline
Alizadeh-2000-v2 
&0.041		&11.14	&8.23	&7.46	&20.51	&1.3  &5.81 &0.09  \\ \hline
Alizadeh-2000-v3
&0.047		&11.4	&8.23	&7.46	&20.26	&1.28 &5.81 &0.09  \\ \hline
Armstrong-2002-v1
&0.062		&10.74	&8.43	&7.56	&11.00 &1.6   &6.91 &0.08  \\ \hline
Armstrong-2002-v2 
&0.065		&27.79	&16.77	&16.67	&20.13  &1.65 &8.75 &0.09 \\ \hline
Bhattacharjee-2001    
&0.189		&153.45	&73.85	&72.61	&21.13 &8.02  &273.79 &0.97\\ \hline
Bittner-2000    
&0.044		&3.07	&1.96	&1.91	&11.28	&1.02 &1.41   &0.05\\  \hline 
Bredel-2005 
&0.033		&2.17	&3.58	&2.94	&15.73	&1.48&0.99    &0.03\\ \hline
Chowdary-2006   
&0.062		&4.86  	&5.67	&3.50	&4.26	&2.69 &31.23  &0.13 \\ \hline
Dyrskjot-2003    
&0.046		&4.55 	&4.83	&4.87	&4.86	&0.81  &4.76  &0.06\\ \hline
Garber-2001
&0.082	&29.97  &18.03	&17.06	&39.47	&1.96   &9.47   &0.11  \\ \hline
Golub-1999-v1
&0.065		&12.13  &8.36	&9.69	&16.33	&1.09 &15.51 &0.13 \\  \hline
Golub-1999-v2
&0.065 &12.13  &8.36	&9.69	&17.42 &1.06  &15.53 &0.13\\  \hline
Gordon-2002    
&0.172		&218.53 &169.01	&158.35	&26.64	&3.57	&543.4 &0.51\\  \hline
Laiho-2002
&0.031		&7.49 	&2.65	&6.73	&4.24	&0.77	&2.27  &0.06\\  \hline
Lapointe-2004-v1 
&0.051		&8.14 	&8.33	&6.05	&18.15	&1.21 &2.40  &0.11 \\  \hline
Lapointe-2004-v2 
&0.092		&36.22   &22.4	&28.82	&18.35	&2.05 &2.56  &0.22 \\  \hline
Liang-2005 
&0.023		&2.36   &1.84	&2.42	&19.57	&0.9 &1.24	&0.06 \\  \hline
Nutt-2003-v1
&0.041		&7.42  	&4.22	&7.51	&6.22 &0.97	&5.91   &0.07 \\  \hline
Nutt-2003-v2
&0.023		&1.08  	&0.45	&1.42	&4.87 &0.67 &1.10   &0.05 \\  \hline
Nutt-2003-v3 
&0.025		&1.61 	&0.53	&2.01	&4.05 &0.55 &0.85   &0.04 \\  \hline
Pomeroy-2002-v1
&0.038		&2.33   &1.1	&3.09	&3.45 &0.57	&1.22   &0.04 \\  \hline
Pomeroy-2002-v2 
&0.031		&3.48   &2.64	&2.58	&13.41	&0.65  &5.19 &0.05	 \\  \hline
Ramaswamy-2001 
&0.397		&136.6	 &188.51	&147.24	&20.92	&3.71 &745.5 &0.58\\  \hline
Risinger-2003 
&0.034		&3.93 	&4.01	&2.31	&6.20   &0.75 &2.19  &0.07 \\  \hline
Shipp-2002-v1    
&0.137		&9.00    &9.44	&7.56	&4.21 &1.04   &15.99 &0.16 \\  \hline
Singh-2002  
&0.084		&4.41   &4.4	&3.89	&4.01 &1.76  &24.3  &0.06 \\  \hline
Su-2001   
&0.436	&166.28  &138.05 &121.82 &22.91  &3.22 &455.1  &0.56\\  \hline
Tomlins  
&0.109	&26.03  &24.6	&22.4 &25.95	&2.44	&20.36 &0.21 \\ \hline
Tomlins-2006-v2 
&0.056	&20.92  &19.33 &17.46 &13.11  &4.02	 &15.14	&0.17 \\ \hline
West-2001  
&0.031	&3.05   &3.81	&2.69&11.25	&1.37 &2.61 &0.06 \\  \hline
Yeoh-2002-v2  
&0.339	&928.1   &784.34 &685.32 &27.97	&6.89 &399.3 &1.33  \\  \hline \hline
\rowcolor{lightgray}
{\bf  Mean Time}  
&0.103	&59.38   &44.35 &50.71	&14.70	&1.61	&82.21 &0.202\\  \hline
\hline 
\rowcolor{lightgray}
{\bf  Median Time}  
&0.062	&8.57   &8.236 &6.76 &14.57	&1.29	&5.86  &0.09\\  \hline
\hline 
\hline
\end{tabular}%
}
\end{table*} 

\bibliography{reference}

\end{document}